\documentclass[letterpaper, 10 pt, conference]{ieeeconf}  

\IEEEoverridecommandlockouts                              

\usepackage{amsmath} 
\usepackage{amssymb}  
\usepackage{subcaption}
\usepackage{cleveref}
\usepackage{arydshln}
\usepackage{graphicx}
 \usepackage{url}
\let\labelindent\relax
\usepackage[inline]{enumitem}

\title{\LARGE \bf
Bridging the Communication Gap: Artificial Agents Learning Sign Language through Imitation
}

\author{Federico Tavella$^{1, 2}$, Aphrodite Galata$^{2}$ and Angelo Cangelosi$^{1, 2}$
\thanks{*Corresponding author
        {\tt\footnotesize federico.tavella@manchester.ac.uk}}
\thanks{$^{1}$Manchester Centre for Robotics and AI, University of Manchester,
        Manchester, M13 9PL, UK}%
\thanks{$^{2}$Department of Computer Science, University of Manchester,
        Manchester, M13 9PL, UK}%
}

\begin{document}

\maketitle
\thispagestyle{empty}
\pagestyle{empty}

\begin{abstract}

Artificial agents, particularly humanoid robots, interact with their environment, objects, and people using cameras, actuators, and physical presence. Their communication methods are often pre-programmed, limiting their actions and interactions. 
Our research explores acquiring non-verbal communication skills through learning from demonstrations, with potential applications in sign language comprehension and expression.
In particular, we focus on imitation learning for artificial agents, exemplified by teaching a simulated humanoid American Sign Language. We use computer vision and deep learning to extract information from videos, and reinforcement learning to enable the agent to replicate observed actions.
Compared to other methods, our approach eliminates the need for additional hardware to acquire information. 
We demonstrate how the combination of these different techniques offers a viable way to learn sign language. Our methodology successfully teaches 5 different signs involving the upper body (i.e., arms and hands). This research paves the way for advanced communication skills in artificial agents.

\end{abstract}

\section{Introduction}

The development of artificial agents like humanoid robots has opened up a wide range of possibilities in various scenarios, spanning from industrial to domestic settings. 
Consequently, communication is crucial for human-robot interaction, whether verbal \cite{mavridisReviewVerbalNonverbal2015} \cite{bonariniCommunicationHumanRobotInteraction2020} or non-verbal \cite{breazealEffectsNonverbalCommunication2005} \cite{mavridisReviewVerbalNonverbal2015} \cite{saundersonHowRobotsInfluence2019} \cite{bonariniCommunicationHumanRobotInteraction2020}. Verbal communication is the default mode for most users, but it is not always suitable or accessible for everyone. Deaf individuals, for instance, rely heavily on non-verbal communication methods such as sign language. This necessitates the development of \textbf{robotic sign language} \cite{tavella2023signs} \cite{loRealizationSignLanguage2016a} \cite{hosseiniTeachingPersianSign2019c} \cite{gagoSequencetoSequenceNaturalLanguage2019} \cite{liangDynamicMovementPrimitive2021a}, where robots are designed to understand and speak sign languages. This field aims to bridge the communication gap between the deaf community and the broader population, allowing seamless interaction and engagement.

According to the World Federation of Deaf ``there are more than 70 million deaf people worldwide'' \cite{un}, while according to WHO ``by 2050 nearly 2.5 billion people are projected to have some degree of hearing loss and at least 700 million will require hearing rehabilitation'' \cite{who}, it is clear that also machines need to learn how to convey messages using a medium different from sound.
The societal benefits of having robots that can speak sign language are extensive and hold immense potential across various fields. 
For example, in the field of education, these robots could assist in classrooms by providing real-time sign language interpretation, enabling deaf students to fully participate and engage in lessons. 

Recently, researchers started approaching the problem of robotics sign language, focusing on imitation \cite{zhangKinematicMotionRetargeting2022} \cite{hosseiniTeachingPersianSign2019c} or translation \cite{gagoSequencetoSequenceNaturalLanguage2019}. 
However, neither of these approaches enables the robot to build an \textit{embodied} representation of different signs based on visual data, as they focus on \textit{retargeting} rather than \textit{learning}. Moreover, they tend to use using additional hardware instead of being vision based. \cite{tavella2023signs} showed how a combination of computer vision and reinforcement learning can be used to learn fingerspelling, but their work does not attempt to address signs involving the full body.

\begin{figure*}
    \centering
  \includegraphics[width=.95\textwidth]{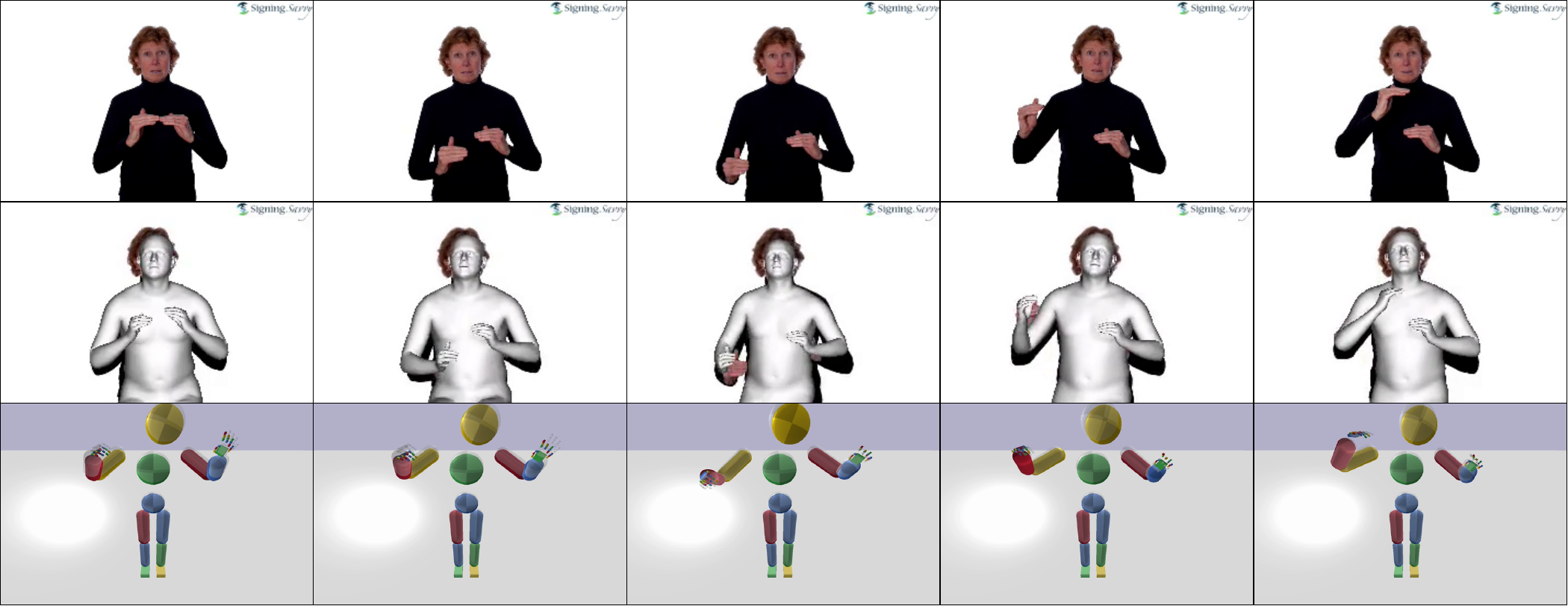}
  \caption{Overview of our proposal. Given a RGB video of a person performing a sign, we use a deep learning approach to extract body information and use such information to teach a simulated humanoid how to perform a specific sign (i.e., \textit{above}).}
  \label{fig:teaser}
\end{figure*}

In conclusion, our work is motivated by a gap in the current state-of-the-art in the field of robotics sign language. In particular, there is no research which approaches the problem of whole body sign language acquisition in artificial agents based on visual data imitation. To summarise, the contributions of our research (illustrated in \Cref{fig:teaser}) are manifold:
\begin{enumerate*}
    \item \textbf{Sign Language Acquisition:} we address the challenging problem of sign language acquisition from RGB video for words, focusing on enabling a machine to imitate sign language gestures which involve both hands and arms;
    \item \textbf{URDF Model Development:} we create a URDF model of a simulated character, which is noteworthy as it is the first known model capable of imitating the whole body and both hands in sign language gestures;
    \item \textbf{Experimental Validation:} Extensive experiments are conducted to optimise parameters for rewards and hyperparameters for training models;
    \item \textbf{Successful Imitation of Signs:} In the end, we achieve a significant milestone by identifying a specific reward structure and set of hyperparameters that enables our approach to successfully learn how to imitate 5 different sign language words, indicating the practical viability of our method.
\end{enumerate*}

\section{Background}

\subsection{Sign language in robotics}\label{subsec:robotics_sl}

Despite all the recent progress, the field of robotics sign language is still very scarce. Most of the research so far focused on robot tutors to aid sign language acquisition in humans \cite{koseHumanoidRobotAssisted2011, koseEvaluationRobotAssisted2012} \cite{koseEffectEmbodimentSign2015} \cite{uluerNewRoboticPlatform2015} \cite{zakipourRASALowCostUpperTorso2016} \cite{scassellatiTeachingLanguageDeaf2018} \cite{zhiTeachingRobotSign2018} \cite{gagoSignLanguageRepresentation2019} \cite{meghdariDesignRealizationSign2019} \cite{luccioLearningSignLanguage2020}. In comparison, very few tried to make robots speak sign language \cite{loRealizationSignLanguage2016a} \cite{hosseiniTeachingPersianSign2019c} \cite{gagoSequencetoSequenceNaturalLanguage2019} \cite{ liangDynamicMovementPrimitive2021a}. Notably, \cite{loRealizationSignLanguage2016a} present an imitation system to teach a humanoid robot to perform sign language by replicating observed demonstrations. By using a Kinect camera and Leap motion sensor, the robot can mimic demonstrated signs based on 3D keypoints and inverse kinematics. Alternatively, in \cite{hosseiniTeachingPersianSign2019c}, the user wears a motion capture suit and performs a sign multiple times to train a set of parallel Hidden Markov Models to encode each sign. Additionally, the authors ensured signs comprehensibility and collision avoidance via a special mapping from the user's workspace to the robot's joint space. The performance of the system was assessed by teaching 10 signs in Persian Sign Language (PSL) to the robot and involving PSL users to investigate how easy was for them to recognise the signs. 8 participants out of ten managed to recognise the signs during the first attempt, while 10 out of 10 did so during the second attempt.
\cite{zhangKinematicMotionRetargeting2022} propose a method for robots to ``learn” sign language by combining learning and optimisation-based motion retargeting. They establish a mapping relationship between the latent space and the robot motion space with a graph decoder. Using the difference between the human and robot motion, they search for the optimal latent code that minimises the gap by gradient descent. Practically, they build an autoencoder and define multiple losses to minimise the difference between human and robot skeletons in terms of joints orientation and end effectors positions.
Hence, despite the advancements in robotic technologies, there remains a glaring lack of research focused on robots that can acquire sign language through imitation, a crucial learning mechanism for humans.

\subsection{Characters animation based on demonstrations}

In the recent literature, we can find several examples of learning-based approaches to control humanoid characters. For instance, \cite{holdenPhasefunctionedNeuralNetworks2017} present a real-time character control mechanism using a novel neural network architecture called a Phase-Functioned Neural Network. In their new approach, they provide as additional input the phase (i.e., a cyclic function like sine or cosine), so that the character can cyclically reproduce the behaviour. For example, walking forward is the repetition of two basic movements in repetition: alternating moving one foot forward after another. In addition to the phase, the authors provide as an input the user's control (e.g., move left, right) and the geometry of the scene to generate motions that reflect the desired control. 
Most notably, \cite{pengDeepMimicExampleguidedDeep2018} \cite{pengSFVReinforcementLearning2018} developed a state-of-the-art approach to skills acquisition. 
In \cite{pengDeepMimicExampleguidedDeep2018}, the authors trained a policy (i.e., a phase-functioned neural network \cite{holdenPhasefunctionedNeuralNetworks2017}) using reinforcement learning based on motion capture data. The motion capture data are acquired from an online library and include skills like running, jumping, doing a backflip, etc. Authors define ad-hoc weighted rewards for different components (e.g., pose, movement velocity, end effectors positions), which indicate to the agent the value of the chosen action. 
Finally, in \cite{pengSFVReinforcementLearning2018}, the authors included a pose estimator and a motion reconstruction component to extract data from RGB videos from YouTube and use them to train the control policy.
\cite{wonScalableApproachControl2020} develop a technique for learning controllers for a large set of different behaviours. They divide the library of motions into clusters of similar motions and train a different policy for each cluster, which combined can reproduce the whole library. 
\cite{pengAMPAdversarialMotion2021} try to obviate the need to manually design rewards and tools for motion selection by using an automated approach based on adversarial imitation learning. The adversarial RL procedure automatically selects which motion to perform, dynamically interpolating and generalising from the dataset.
Finally, \cite{pengASELargeScaleReusable2022} combine techniques from adversarial imitation learning and unsupervised reinforcement learning to develop skill embeddings that produce life-like behaviours. Most interestingly, they achieve significant speed-up by parallelising the training by leveraging a novel GPU-based simulator, decreasing the extensive time required to train a single policy.
Similarly to our proposal, \cite{tavella2023signs} focuses on teaching robots sign language fingerspelling through imitation learning from RGB videos. They develop a model for a robotic hand and compare two learning algorithms (PPO \cite{schulmanProximalPolicyOptimization2017} and SAC \cite{haarnoja2018soft}). The study successfully imitates six different fingerspelled letters with performance similar to ideal retargeting.

\section{Methods}

We begin by discussing how we construct our whole body model by combining existing body and hand models. We then provide a brief overview of how we use a deep learning model to analyse RGB videos and extract data necessary for learning to imitate -- namely, joint rotations. Next, we describe how we use reinforcement learning to imitate sign language based on the information extracted from the videos.

\subsection{Whole body model}

\begin{figure}[ht]
    \centering
    \begin{subfigure}{.15\textwidth}
        \centering
        \includegraphics[width=\textwidth]{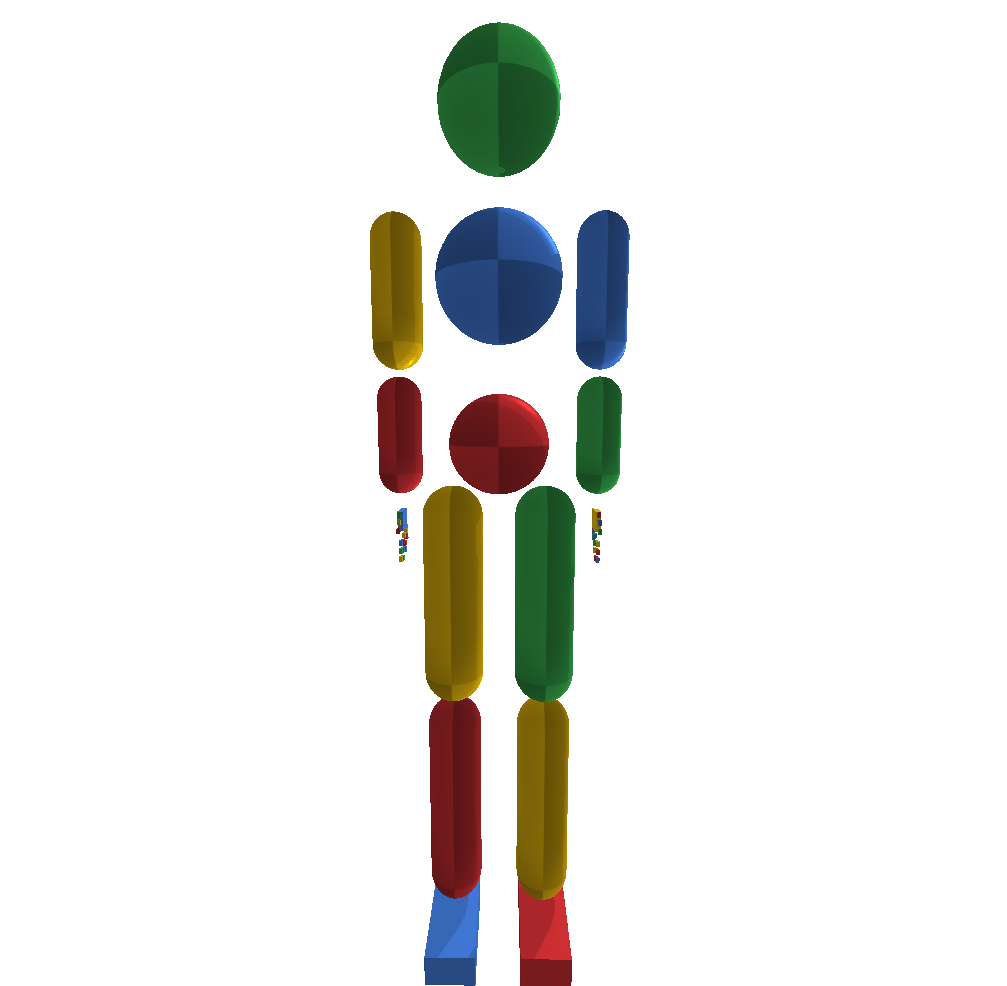}
        \caption{Front view}
        \label{fig:whole_urdf_front}
    \end{subfigure}
    \hfill
    \begin{subfigure}{.15\textwidth}
        \centering
        \includegraphics[width=\textwidth]{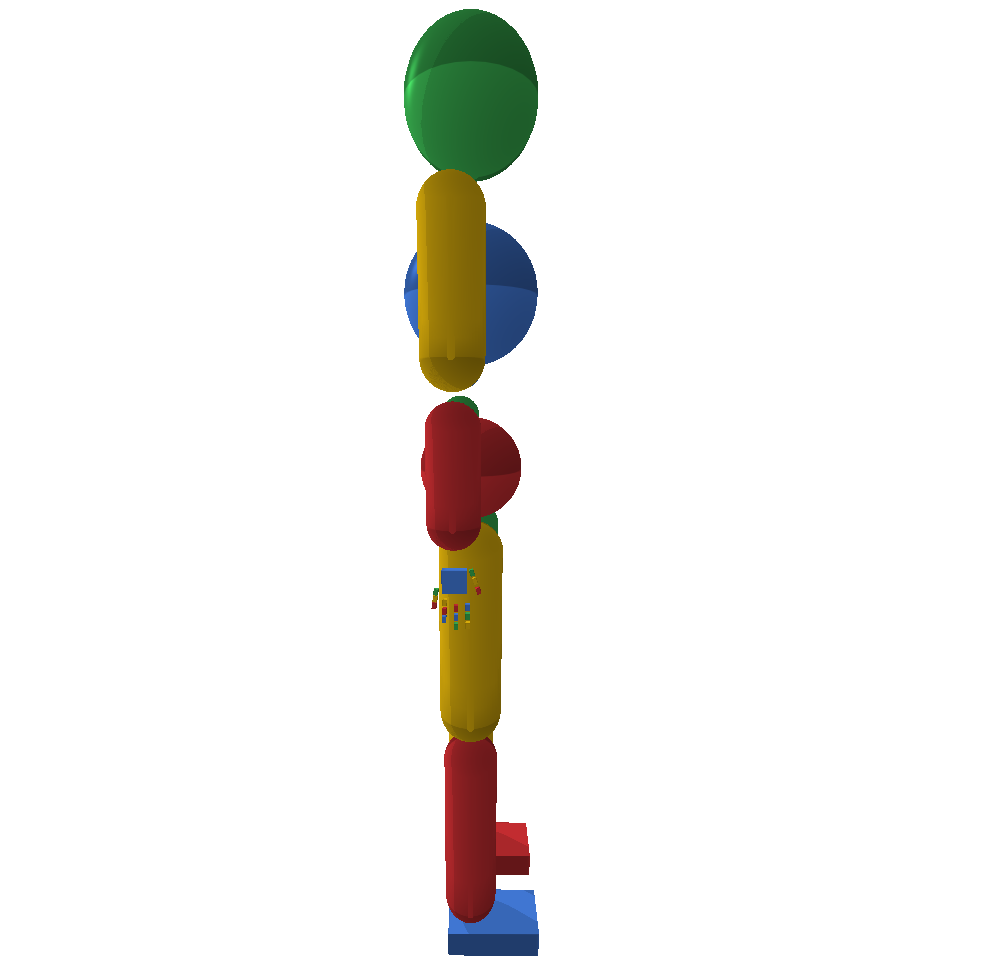}
        \caption{Side view}
        \label{fig:whole_urdf_side}
    \end{subfigure}
    \hfill
    \begin{subfigure}{.15\textwidth}
        \centering
        \includegraphics[width=\textwidth]{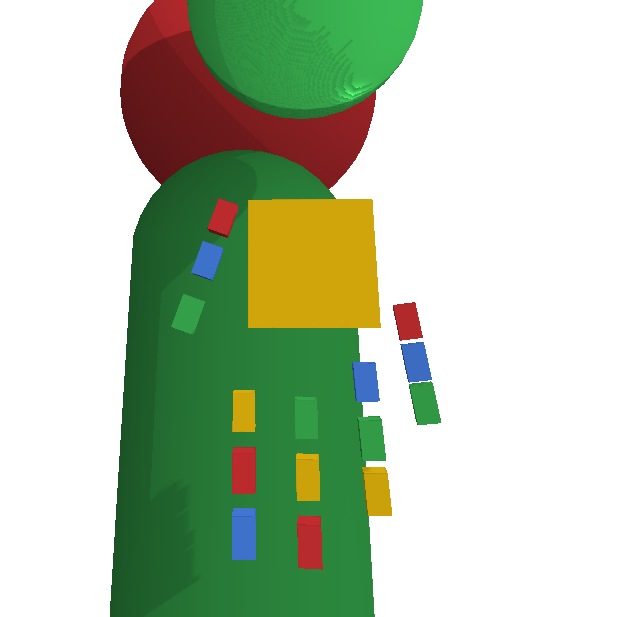}
        \caption{Hand zoom in}
        \label{fig:whole_urdf_hand}
    \end{subfigure}
    \caption{Whole body model. We integrated an available body model \cite{pengDeepMimicExampleguidedDeep2018} with a hand model \cite{tavella2023signs}, replicated and mirrored to obtain both left and right hands.}
    \label{fig:whole_urdf}
\end{figure}

We re-adapt a humanoid model available in the literature \cite{pengDeepMimicExampleguidedDeep2018} to our scenario. 
It is composed of 13 joints (2 shoulder, elbows, hips, knees, ankles and 1 neck, chest and waist/root). 9 out of 13 joints have 3 degrees of freedom, while the remaining 4 (2 knees and 2 elbows) have only 1 Degree of Freedom (DoF), for a total of 31 DoFs. It is 1.62cm tall and weighs 45 kg. As for the hands, we reuse the model we describe in \cite{tavella2023signs}, but we duplicate the model and mirror it to reflect the orientation of the right and left hands. Moreover, we change the shape of the wrist from a small cube to a parallelepiped in order to make it resemble the palm of a hand. \Cref{fig:whole_urdf} shows our final model from different points of view.


We simulate the humanoid using PyBullet \cite{pybullet}, an open-source physics simulator. For each joint motor, the simulated controller calculates the error as $\varepsilon = k_p \Delta P + k_d \Delta V$, where $k_p$ and $k_d$ are, respectively, the position and velocity gains, and $\Delta P = P - \hat{P}$ and $\Delta V = V - \hat{V}$ are the position and velocity errors (i.e., the difference between the desired and the actual value respectively).

\subsection{Motion extraction and imitation}

\begin{figure}[ht]
    \centering
    \begin{subfigure}{.22\textwidth}
        \centering
        \includegraphics[width=\textwidth]{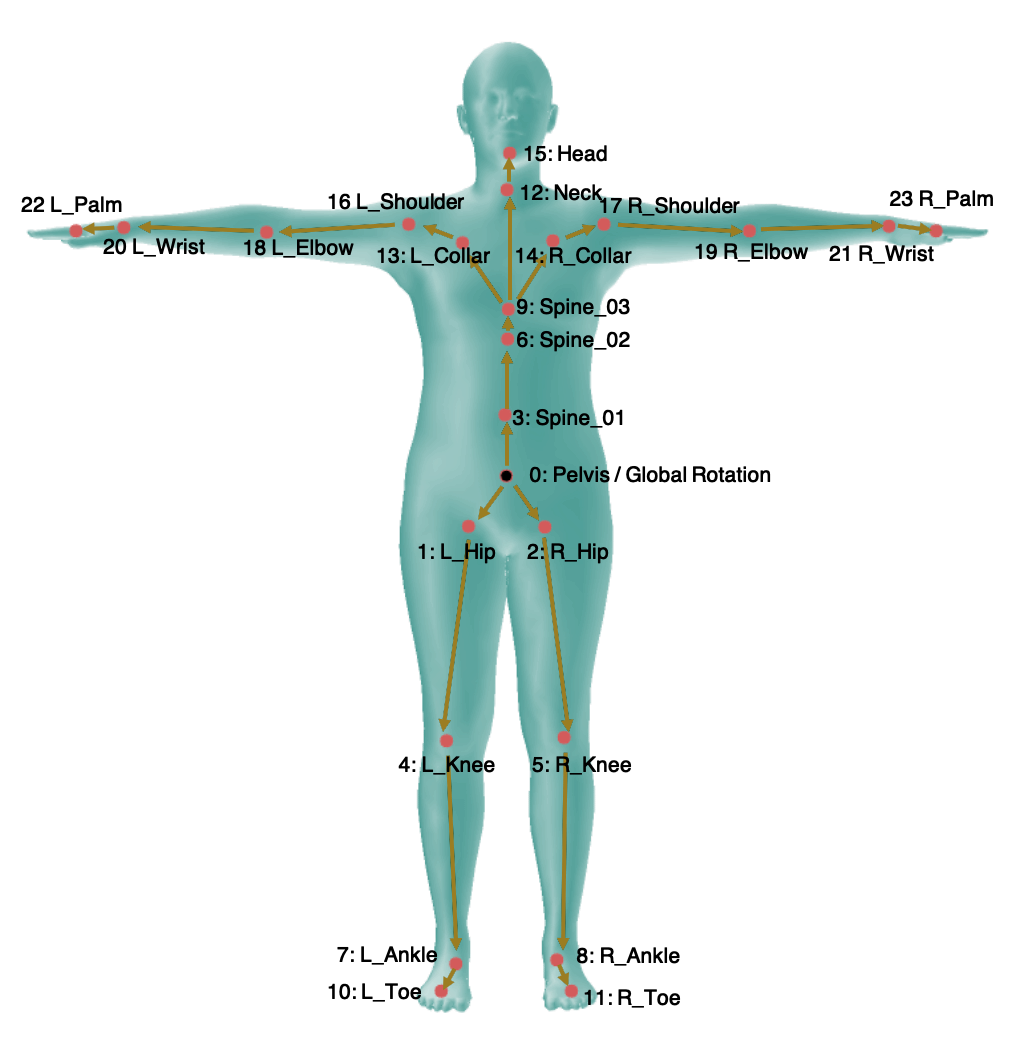}
        \caption{Body joints}
        \label{fig:frank_body}
    \end{subfigure}
    \hfill
    \begin{subfigure}{.25\textwidth}
        \centering
        \includegraphics[width=.7\textwidth]{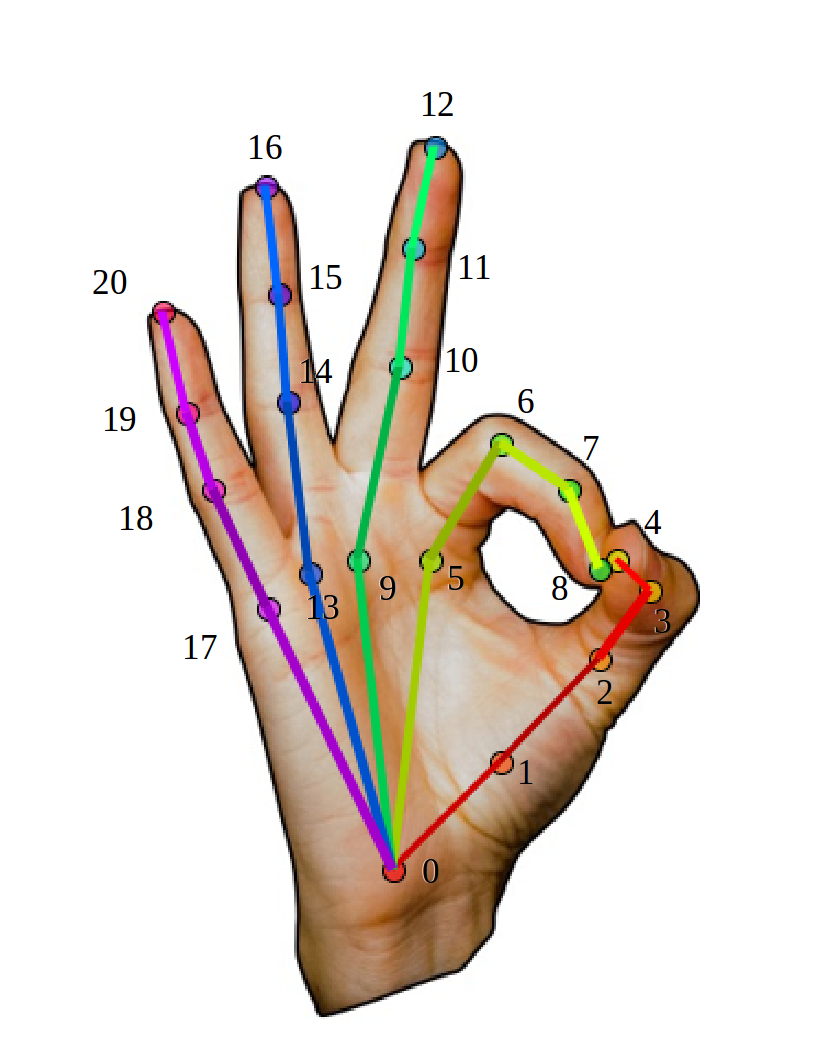}
        \caption{Hand joints}
        \label{fig:frank_hand}
    \end{subfigure}
    \caption{Body and hand keypoints extracted using FrankMocap. The body is composed of 24 keypoints, while each hand has 21 keypoints. (Both reprinted from \cite{frankimg})}
    \label{fig:frank_whole}
\end{figure}

The work by Peng et al. \cite{pengSFVReinforcementLearning2018} employs Human Motion Reconstruction \cite{kanazawaEndtoEndRecoveryHuman2018} in conjunction with a motion reconstruction method to derive rotational information from videos, subsequently feeding it into DeepMimic \cite{pengDeepMimicExampleguidedDeep2018}. It should be noted that their study focused on specific, unconventional movements such as backflips, which are typically absent from datasets used to train such models. In contrast, our particular context involves more general motions, prompting us to utilise FrankMocap \cite{rongFrankMocapMonocular3D2021} to extract 3D rotations and keypoints for both the upper body and hands. 
Both models were used in the literature to extract information from sign language videos \cite{tavellaPhonologyRecognitionAmerican2022} \cite{tavellaWLASLLEXDatasetRecognising2022}.
\Cref{fig:frank_whole} provides a visual depiction of the extracted keypoints and rotations for the upper body and hands. However, it is important to highlight that, given our emphasis on sign language, we do not consider information related to the lower body, specifically the legs.

Lately, imitation learning and behavioural cloning \cite{suttonReinforcementLearningSecond2018} took advantage of many techniques from Reinforcement Learning (RL). RL is commonly represented as a Markov Decision Process (MDP) \cite{bellmanMarkovianDecisionProcess1957}, which comprises four key components: a set of states $S$, a set of actions $A$, a reward function $R$, and the transition probability $P_a(s, s')$. An agent engaged in learning observes the environment and its current state $s$. Based on these observations (i.e., the state), it selects an action $a$ to transition to a new state $s'$ according to the probability given by
$
P_{a}(s, s') = P(s_{t+1} = s' | s_t = s, a_t = a).
$
This transition yields a reward $R_{a}(s, s')$, or $r$, indicating the desirability of the chosen action. The ultimate goal is to acquire a policy, a mapping from states to actions, that maximises the expected cumulative reward. Additionally, a policy can be parametric, denoted as $\pi_{\theta}$. In this context, the policy needs to identify the optimal parameters $\theta^*$ that maximise the expected cumulative reward.

Various algorithms are available for discovering an optimal policy, with the selection typically influenced by the nature of the action and/or state spaces (discrete or continuous). For more comprehensive information, readers are directed to \cite{suttonReinforcementLearningSecond2018}. In our specific case, we employ the Proximal Policy Optimization (PPO) algorithm \cite{schulmanProximalPolicyOptimization2017} to estimate a policy for our problem.

PPO is an optimisation algorithm used when both action and state spaces are continuous. It is a policy gradient method, where the gradient of the expected cumulative reward is calculated using trajectories $\tau$ -- i.e., sequences of $(s, a, r)$ over a set of contiguous time steps -- sampled by following the policy. Thus, given a parametric policy $\pi_{\theta}$ and $T$ steps, the expected reward is

\begin{equation}
    J(\theta) = \mathbb{E}_{\tau \sim p_{\theta}(\tau)} \left[ \sum_{t=0}^{T} \gamma^t r_t \right]
\end{equation}

where 
$p_{\theta}(\tau) = p(s_0) \prod_{t=0}^{T-1} p(s_{t+1} | s_t, a_t) \pi_{\theta}(a_t | s_t)$
is the distribution over all possible trajectories 
$\tau = (s_0, a_0, s_1, a_1, ..., a_{T-1}, s_{T})$
induced by the policy $\pi_\theta$, $p(s_0)$ being the initial state distribution and $\gamma \in [0, 1]$ is a discount factor used to ensure that the reward has an upper bound. The policy gradient can be estimated as
$\nabla_\theta J(\theta) = \mathbb{E} \left[ \nabla_\theta log(\pi_\theta(a_t | s_t)) \mathcal{A}_t \right]$
where $\mathcal{A}_t = R_t - V(s_t)$ represents the advantage of taking an action $a_t$ from a given state $s_t$, $R_t$ is the cumulative reward by a particular trajectory starting from state $s_t$ at time t, and 
$
V(s_t) = \mathbb{E}_t \left[ R_t | \pi_\theta, s_t \right]
$
is the value function that estimates the average reward for starting at $s_t$ and following the policy for all subsequent steps.

We formulate the sign language acquisition problem as a control problem. We model it as an MDP and use PPO \cite{schulmanProximalPolicyOptimization2017} to learn a policy for each different motion. However, as we added a full body model, it was not clear whether it could be possible to reuse the scaling factors from \cite{pengDeepMimicExampleguidedDeep2018} and \cite{tavella2023signs} for the subrewards. One option would be to recalibrate the weights of each different sub-reward, but that would imply additional experiments to find the optimal balance between different rewards. In a resource eager scenario like when using reinforcement learning, additional experiments are a major drawback.
Hence, we take inspiration from \cite{wonScalableApproachControl2020} and define our reward as a \textit{multiplicative} reward (rather than an additive) as
$
r_t = r_t^p \cdot r_t^v \cdot r_t^e \cdot r_t^r
$
where $r_t^x$ is the reward at time $t$ for quantity $x$ (which can be body, velocity, end effectors and root), calculated as
$r_t^x = e^{-k^x \varepsilon_t^x}$
with $\varepsilon_t^x$ being the error for quantity $x$ at time $t$, and $k^x$ being a factor used to balance the reward based on the scale of the error, as described in \cite{pengDeepMimicExampleguidedDeep2018}. In addition, we redefine $r_t^p$ and $r_t^v$ as
$
r_t^p = r_t^{p, b} \cdot r_t^{p, h}
$
and
$
r_t^v = r_t^{v, b} \cdot r_t^{v, h},
$
with $b$ indicating the body and $h$ the hands. We choose to divide the rewards (and the errors) for body and hands as we believe this give us a more clear indication of what the algorithm can (not) learn. As for the end-effectors reward $r_t^e$, we consider just the position of the wrists, as opposed to \cite{tavella2023signs} in which they considered the position of the fingertips.
\Cref{tab:compare} summarises the differences between the three approaches. Complementary, scaling factors are the same from \cite{pengDeepMimicExampleguidedDeep2018}, for both body and hands.

\begin{table}[ht]
\centering
\begin{tabular}{lccc}
\hline \hline
                     & DeepMimic & SiLa & Ours \\ \hline
Number of joints     & 13        & 16       & 45         \\
Number of DoFs       & 34        & 15       & 50         \\
State space size     & 197       & 210      & 509        \\
Action space size    & 36        & 15       & 57         \\ \hline \hline
\end{tabular}%
\caption{Comparison between DeepMimic \cite{pengDeepMimicExampleguidedDeep2018}, SiLa \cite{tavella2023signs} for fingerspelling and our whole body approach.}
\label{tab:compare}
\end{table}
\section{Experimental setup}

\subsection{Controller tuning}

As for the controllers tuning, we do not need to go through the whole procedure described in \cite{tavella2023signs} for the simulated hand, as we can reuse the values of the controllers pose gain $k_p$ and velocity gain $k_d$ from \cite{pengDeepMimicExampleguidedDeep2018} for the body and from \cite{tavella2023signs} for the hands. However, we need to refine the values for the arms, as the fact that we attached end effectors to them means that we also changed the dynamics of the whole arms. As opposed to \cite{tavella2023signs} approach to tuning the controller parameters, we do not deploy a completely automated approach, but rather an empirical one. Considering we aim to adapt only the values for 2 joints (i.e., shoulder and elbow), we find it easier to manually adapt the values balancing the body and hand error rather than develop a sophisticated function to find an acceptable equilibrium between the two errors.

\subsection{Motion imitation}

We exploit a motion file from \cite{rongFrankMocapMonocular3D2021} for the hyperparameter tuning of the multilayer perceptron representing our policy. However, preliminary results immediately showed that the scaling factors for the hands (that were the same as for \cite{pengDeepMimicExampleguidedDeep2018}) did not yield satisfactory results. In fact, as the reward is a product of sub-rewards, if one of the sub-rewards is 0, then the total reward will be 0 as well. Thus, we run additional searches for the parameter for hand pose and velocity. We use a mixture of automated and manual approaches, alternating hyperparameters and scaling factors explorations. We test the permutation of the following values for pose and velocity: [2, 1, 0.5, 0.2] and [$10^{-1}$, $10^{-2}$, $10^{-3}$, $10^{-4}$] respectively.
Intuitively, given that the reward $r_t^x = e^{-k^x \varepsilon_t^x}$, tuning the scaling factor $k^x$ makes the reward more (or less) lenient towards the error $x$. 



We perform a hyperparameter search based on Bayesian optimisation to find the ideal hyperparameters. When performing such a search, we train our algorithm for 25 million steps, which is 50\% of how long we train to achieve imitation for the final policies. The set over which we perform the search is summarised in \Cref{tab:hyperparams_whole}. 
Finally, we test the generalisability of such hyperparameters over five different signs, repeating each attempt 10 times using different seeds and training for 50 million steps. We choose the different signs via a qualitative evaluation of the output of FrankMocap. 
We choose a small amount of different motions to imitate due to extensive time required to train and statistically validate multiple policies.
In addition, when choosing the signs we also take into account the limitations of the model from \cite{tavella2023signs} (e.g., only 1 DoF for the elbow, wrist or phalanges of each finger). The lemmas corresponding to the five signs are \texttt{above}, \texttt{snow}, \texttt{father}, \texttt{mother} and \texttt{yes}. Their respective identifiers from WLASL \cite{caselliASLLEXLexicalDatabase2017} are \texttt{00433}, \texttt{52861}, \texttt{69318}, \texttt{69402} and \texttt{69546}, which we use as labels in our results.

\begin{table}[ht]
\centering
\begin{tabular}{ll}
\hline \hline
\textbf{Parameter}    & \textbf{Values}                        \\ \hline
hidden layers size              & (512-1024-512), (256-512-256) \\
learning rate                   & (1, 3, 10, 30, 100) x $10^{-6}$  \\
number of steps                 & 512, 1024, 4096                 \\
batch size                      & 128, 256, 512                 \\
log std dev                     & -5, -3, -2, -1                             \\
discount factor                 & 0.9, 0.95           \\
number of epochs                & 3, 5, 10                 \\ \hline \hline
\end{tabular}
\caption{Values of different hyperparameters for PPO.}
\label{tab:hyperparams_whole}
\end{table}
\section{Results}

Our fist steps aim at finding the ideal scaling factors $k^{p, h}$ and $k^{v, h}$ for the rewards $r_t^{p,h}$ and $r_t^{v,h}$.
We explore the permutations of $k^{p, h}$ = [2, 1, 0.5, 0.2] and $k^{v, h}$ = [$10^{-1}$, $10^{-2}$, $10^{-3}$, $10^{-4}$] using a Bayesian approach. 
We find two combinations of parameters which provide the best results: one with $k^{p, h} = 0.5$ and $k^{v, h} = 5 \cdot 10^{-4}$, and one with $k^{p, h} = 0.2$ and $k^{v, h} = 10^{-4}$. This means that, in order to enable our algorithm to be able to learn, our reward needs to be more lenient with the errors regarding the pose and velocity of the hands, as the previous values used in \cite{tavella2023signs} were $k^{p, h} = 0.5$ and $k^{v, h} = 0.05$.
Consequently, we perform two hyperparameters searches to test different combinations of values $k^{p, h}$ and $k^{v, h}$. 
In particular, we test $k^{p, h} = 0.5$ and $k^{v, h}= 5 \cdot 10^{-4}$ versus $k^{p, h} = 0.2$ and $k^{v, h}=10^{-4}$.
The former does not provide any satisfactory result (i.e., max reward after 25 million steps around 0.02), while the latter provides some interesting candidates, as a couple of runs have a reward higher than 0.4 at 50\% of the training. Thus, we confirm that the ideal parameters are $k^{p, h} = 0.2$ and $k^{v, h}=10^{-4}$. 

\begin{figure}[ht]
    \centering
    \begin{subfigure}{.35\textwidth}
        \centering
        \includegraphics[width=\textwidth]{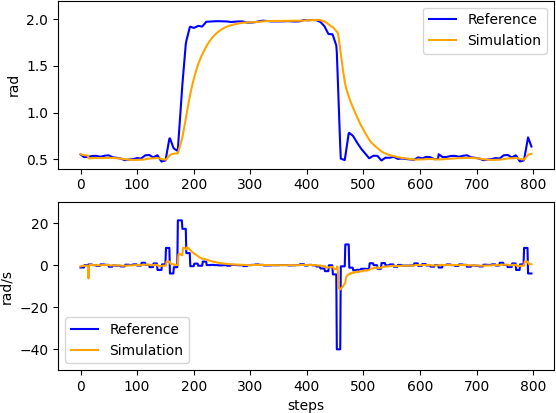}
        \label{fig:old_elbow}
    \end{subfigure}
    \vfill
    \begin{subfigure}{.35\textwidth}
        \centering
        \includegraphics[width=\textwidth]{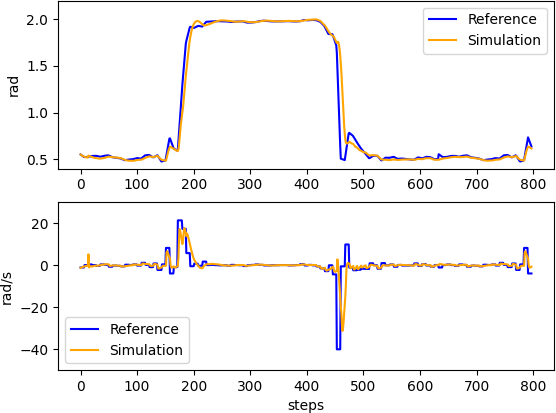}
        \label{fig:new_elbow}
    \end{subfigure}
    \caption{Elbow dynamics with $k_d = 40$ (top) vs $8$ (bottom).}
    \label{fig:elbow}
\end{figure}

We test these parameters on the sign \texttt{above}. We analyse the results by checking the error of the imitation results -- i.e., the difference between the reference motion and the simulated one. By doing so, we observe a delay between the replicated and reference position of elbows. Thus, we modify the velocity gain $k_d$ for the shoulder and elbow, in order to improve the responsiveness of our model. By changing the value from 40 to 8 for the shoulder and 30 to 6 for the elbow, we notice a significant improvement (\Cref{fig:elbow}) without significantly affecting the rest of the connected joints, like the wrist.

In addition, as we changed one of the parameters of the controller and slightly increased the velocity error, we decreased the scaling factor via some manual tuning. We change $k^{v, b}$ from $10^{-1}$ to $5 \cdot 10^{-3}$ so that the reward related to the body velocity error is higher. In the end, we select the configuration summarised in \Cref{tab:hyperparams_first}.

\begin{table}[ht]
\centering
\begin{tabular}{|cc|c|c|}
\hline
\multicolumn{2}{|c|}{\textbf{hidden layers size}}                & \textbf{learning rate} & \textbf{N steps}  \\ \hline
\multicolumn{2}{|c|}{256, 512, 256}                              & $3 \cdot 10^{-6}$      & 1024              \\ \hline
\multicolumn{1}{|c|}{\textbf{batch size}} & \textbf{log std dev} & \textbf{discount}      & \textbf{N epochs} \\ \hline
\multicolumn{1}{|c|}{128}                 & -3                   & 0.95                   & 10                \\ \hline
\end{tabular}%
\caption{Final set of hyperparameters selected using the tuning motion.}
\label{tab:hyperparams_first}
\end{table}

\Cref{tab:scale_rews_tune} summarises the different scaling factors we explore. In addition, we provide an approximation of the component $r^p \cdot r^v$ of the reward, based on the error measured while replicating the original movement. We can see how changing $k^{b, v}$ can bring a (hypothetical) increment of approximately 10\%. Given that we identified reward and hyperparameters, we proceed to test the generalisability of such values over different motions. 

\begin{table}[ht]
\centering
\begin{tabular}{ccccc}
\\ \hline \hline
                         & \textbf{Default} & \textbf{Run 1}    & \textbf{Run 2}    & \textbf{Run 3}    \\ \hline
\textbf{$k^{p, b}$}      & 2                & 2                 & 2                 & \textbf{2}        \\
\textbf{$k^{p, h}$}      & 2                & $5 \cdot 10^{-1}$ & $2 \cdot 10^{-1}$ & $2 \cdot 10^{-1}$ \\
\textbf{$k^{v, b}$}      & $10^{-1}$        & $10^{-1}$         & $10^{-1}$         & $5 \cdot 10^{-3}$ \\
\textbf{$k^{v, h}$}      & $10^{-1}$        & $5 \cdot 10^{-4}$ & $10^{-4}$         & $10^{-4}$         \\ \hdashline
\textbf{$r^p \cdot r^v$} & 0                & 0.297             & 0.694             & \textbf{0.790}             \\ \hline \hline
\end{tabular}%
\caption{Estimated sub-rewards for different scaling values, calculate using the tuning motion.}
\label{tab:scale_rews_tune}
\end{table}

\begin{figure}[ht]
    \centering
    \includegraphics[width=.45\textwidth]{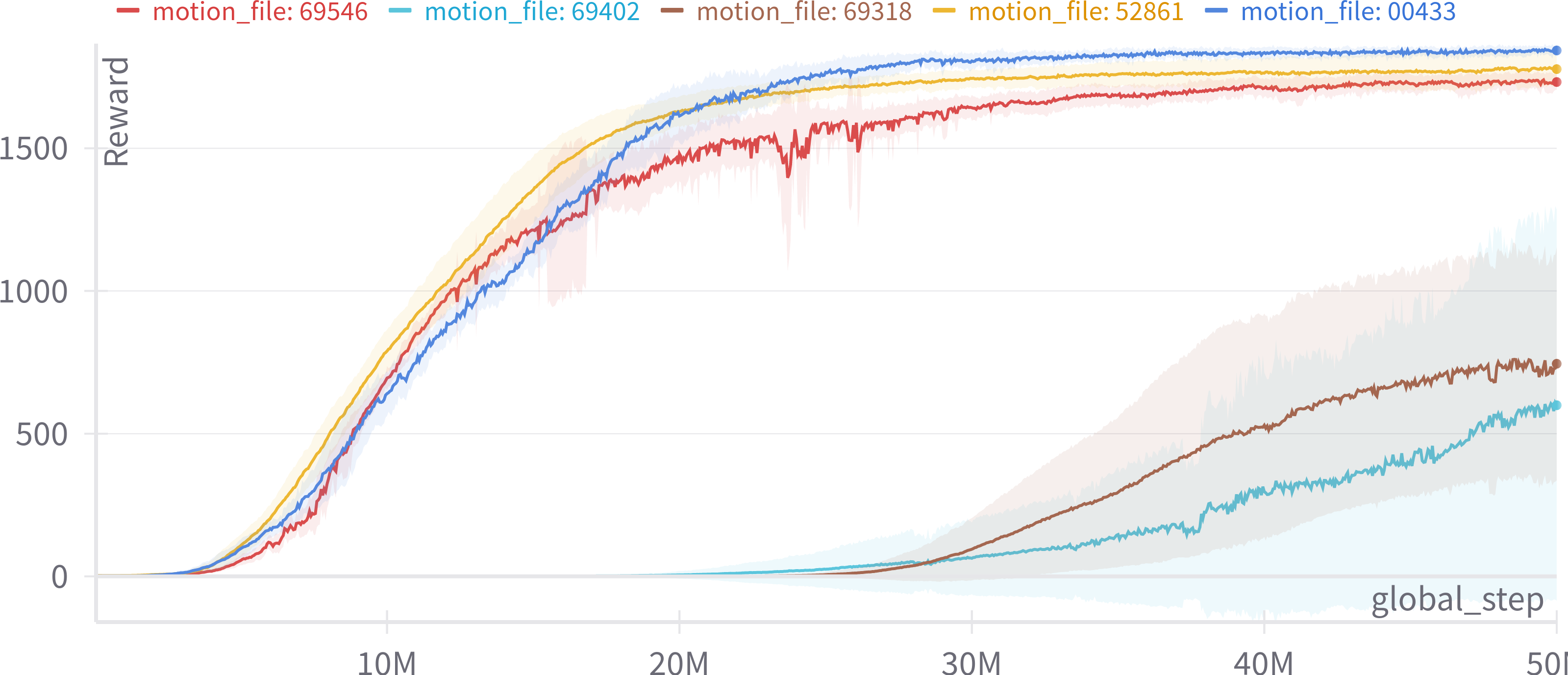}
    \caption{First run of 5 signs, the average cumulative reward is calculated over 10 different seeds}
    \label{fig:first_final}
\end{figure}

\Cref{fig:first_final} shows the results for the five different signs. Each sign is calculated as mean (and standard deviation) over 10 different seeds. We can see that, out of five different signs, only three achieve a cumulative reward of around 1800. The two lemmas for which the algorithm does not converge to acceptable results are \texttt{father} and \texttt{mother}. The signs are similar, as both involve placing the hand fully open, with the former attaching the thumb forehead and the latter to the chin. 

By looking at the single runs, we notice that the error never converges over 7 different runs with different seeds.
Considering we change the reference motion, we roll back to test two previous configurations for the scaling factors. In this way, we can double-check whether $k_{p,h}$ and $k_{v, b}$ actually contribute to the learning process. We carry out a hyperparameters search over the following configurations, summarised by the first two columns of \Cref{tab:scale_rew_father}. 


\begin{table}[ht]
\centering
\begin{tabular}{ccccc}
\\ \hline \hline
                & \textbf{Run 1}    & \textbf{Run 2}    & \textbf{Run 3}    & \textbf{Run 4}    \\ \hline
\textbf{$k^{p, b}$}      & 2                 & 2                 & 1                 & $5 \cdot 10^{-1}$ \\
\textbf{$k^{p, h}$}      & 2                 & $2 \cdot 10^{-1}$ & $2 \cdot 10^{-1}$ & $2 \cdot 10^{-1}$ \\
\textbf{$k^{v, b}$}      & $5 \cdot 10^{-3}$ & $10^{-1}$         & $10^{-3}$         & $5 \cdot 10^{-3}$ \\
\textbf{$k^{v, h}$}      & $10^{-4}$         & $5 \cdot 10^{-4}$ & $10^{-4}$         & $10^{-4}$         \\ \hdashline
\textbf{$r^p \cdot r^v$} & 0.656             & 0.060             & \textbf{0.785}             & 0.715   \\ \hline \hline         
\end{tabular}%
\caption{Estimated sub-rewards for different scaling values, calculate using the \texttt{father} motion.}
\label{tab:scale_rew_father}
\end{table}
 
The first one aims to reduce the relevance of the reward regarding of the pose of the hands, while the second one targets the body and hand velocity. However,
neither of the sweeps produces a single run in which the different error components converge, and thus the reward is approximately 0.
Hence, we try to modify the scaling factors according to the last 2 columns of \Cref{tab:scale_rew_father}. In the first run, we change all the factors except for the one associated with the pose of the hands. We can see a significant improvement over the previous sweep. However, we notice that the reward for the body pose oscillates considerably, while the reward for the body velocity has a high value but with very few variations. Thus, we try to address these issues with our second sweep. We reduce the scaling factor associated with the body pose and increase the one associated with the body velocity. From this final sweep, we select a set of hyperparameters to test over the other signs. \Cref{tab:hyperparams_second} summarises the best set of hyperparameters we discover in our extensive exploration. Overall, the only difference is the parameters regarding the number of epochs and the number of steps, which changed from 10 to 5 and 1024 to 512 respectively.


\begin{table}[ht]
\centering
\begin{tabular}{|cc|c|c|}
\hline
\multicolumn{2}{|c|}{\textbf{hidden layers size}}                & \textbf{learning rate} & \textbf{N steps}  \\ \hline
\multicolumn{2}{|c|}{256, 512, 256}                              & $3 \cdot 10^{-6}$      & 512               \\ \hline
\multicolumn{1}{|c|}{\textbf{batch size}} & \textbf{log std dev} & \textbf{discount}      & \textbf{N epochs} \\ \hline
\multicolumn{1}{|c|}{128}                 & -3                   & 0.95                   & 5                 \\ \hline
\end{tabular}%
\caption{Selected hyperparameters using the motion representing the sign \texttt{father}. These hyperparameters are selected following quantitative (i.e., reward) and qualitative (i.e., visually analysing the training curve) evaluations.}
\label{tab:hyperparams_second}
\end{table}

\begin{figure}[ht]
    \centering
    \includegraphics[width=.45\textwidth]{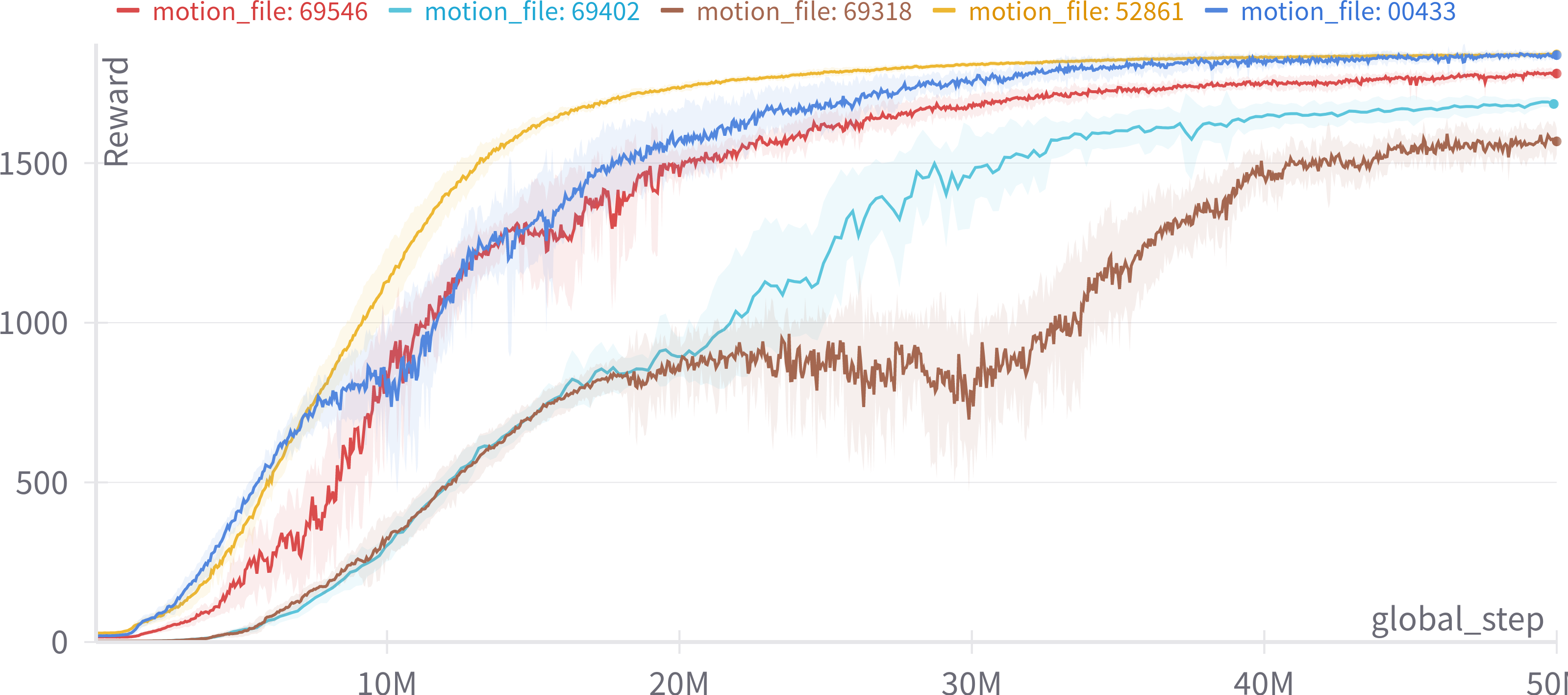}
    \caption{Final run, the average cumulative reward is calculated over 10 different seeds}
    \label{fig:second_final}
\end{figure}

Finally, we train our model with the selected hyperparameters over the different signs. \Cref{fig:second_final} shows the learning curves as average (and standard deviation) over 10 different seeds. We can see how, when compared to \Cref{fig:first_final}, the learning curve is less steep. However, all five signs achieve a cumulative reward above 1500, as opposed to our previous attempt in which two of them did not surpass 750. \Cref{tab:final_comparison} provides a convenient comparison of the results of the two different runs and the performance ceiling of ideal retargeting.

\begin{table}[ht]
\centering
\begin{tabular}{cccc}
\hline
\textbf{Motion} & \textbf{Run 1}      & \textbf{Run 2}      & \textbf{Retargeting} \\ \hline
00433           & \underline{$1843 \pm 16$} & $1838 \pm 13$       & 1976                 \\
52861           & $1777 \pm 62$       & \underline{ $1840 \pm 9$}  & 1961                 \\
69318           & $745 \pm 408$       & \underline{ $1568 \pm 41$} & 1954                 \\
69402           & $600 \pm 685$       & \underline{ $1685 \pm 16$} & 1961                 \\
69546           & $1732 \pm 27$       & \underline{ $1780 \pm 16$} & 1967                 \\ \hline
\end{tabular}%
\caption{Cumulative rewards over 2000 steps, comparing the two different sets of hyperparameters and ideal retargeting}
\label{tab:final_comparison}
\end{table}
\section{Discussion}

Our paper introduces a novel approach to address the problem of sign language acquisition from demonstration. 
We have successfully developed a URDF model of a humanoid with two hands, each equipped with dexterous fingers. Our model stands as a novel solution enabling the simultaneous imitation of body and hand movements. Leveraging off-the-shelf pre-trained pose estimation models, such as FrankMocap \cite{rongFrankMocapMonocular3D2021}, we eliminate the need for additional hardware to capture spatio-temporal information. Furthermore, we elucidate the methodology for modelling rewards, facilitating imitation via a well-established reinforcement learning algorithm, such as PPO. Our extensive experimentation on reward parameters and model hyperparameters illustrates the iterative process required when using reinforcement learning. We identified an optimal set of parameters that enables our approach to effectively imitate five different sign language signs. Consequently, we can assert that our proposal offers a viable avenue for acquiring sign language.

Looking ahead, our work opens up to several compelling future directions. Firstly, enhancing our approach involves increasing the DoFs for each hand joint from 1 to 2, addressing a limitation identified in the model provided by \cite{tavella2023signs}. This enhancement would broaden the range of sign language signs that our system can effectively imitate. 
Another avenue for improvement lies in devising mechanisms for leveraging previous experiences to expedite the learning process for new signs.
Finally, the ultimate frontier in deploying our research lies in the transition from a simulated environment to a real-world setting. This would entail the integration of our policies into a physical robot capable of performing signs, with the added challenge of human recognition of the signs executed by the robot. This would be a significant milestone in making sign language more accessible to individuals who rely on it as their primary mode of communication.



\bibliographystyle{IEEEtran}
\bibliography{ml,sign_language,robotics,psy,animation,extra}

\begin{thebibliography}{10}
\providecommand{\url}[1]{#1}
\csname url@rmstyle\endcsname
\providecommand{\newblock}{\relax}
\providecommand{\bibinfo}[2]{#2}
\providecommand\BIBentrySTDinterwordspacing{\spaceskip=0pt\relax}
\providecommand\BIBentryALTinterwordstretchfactor{4}
\providecommand\BIBentryALTinterwordspacing{\spaceskip=\fontdimen2\font plus
\BIBentryALTinterwordstretchfactor\fontdimen3\font minus \fontdimen4\font\relax}
\providecommand\BIBforeignlanguage[2]{{%
\expandafter\ifx\csname l@#1\endcsname\relax
\typeout{** WARNING: IEEEtran.bst: No hyphenation pattern has been}%
\typeout{** loaded for the language `#1'. Using the pattern for}%
\typeout{** the default language instead.}%
\else
\language=\csname l@#1\endcsname
\fi
#2}}

\bibitem{mavridisReviewVerbalNonverbal2015}
N.~Mavridis, ``A review of verbal and non-verbal human\textendash robot interactive communication,'' \emph{Robotics and Autonomous Systems}, vol.~63, pp. 22--35, Jan. 2015.

\bibitem{bonariniCommunicationHumanRobotInteraction2020}
A.~Bonarini, ``Communication in {{Human-Robot Interaction}},'' \emph{Current Robotics Reports}, vol.~1, no.~4, pp. 279--285, Dec. 2020.

\bibitem{breazealEffectsNonverbalCommunication2005}
C.~Breazeal, C.~Kidd, A.~Thomaz, G.~Hoffman, and M.~Berlin, ``Effects of nonverbal communication on efficiency and robustness in human-robot teamwork,'' in \emph{2005 {{IEEE}}/{{RSJ International Conference}} on {{Intelligent Robots}} and {{Systems}}}, Aug. 2005, pp. 708--713.

\bibitem{saundersonHowRobotsInfluence2019}
S.~Saunderson and G.~Nejat, ``How {{Robots Influence Humans}}: {{A Survey}} of {{Nonverbal Communication}} in {{Social Human}}\textendash{{Robot Interaction}},'' \emph{International Journal of Social Robotics}, vol.~11, no.~4, pp. 575--608, Aug. 2019.

\bibitem{tavella2023signs}
F.~Tavella, A.~Galata, and A.~Cangelosi, ``Signs of language: Embodied sign language fingerspelling acquisition from demonstrations for human-robot interaction,'' 2023.

\bibitem{loRealizationSignLanguage2016a}
S.-Y. Lo and H.-P. Huang, ``Realization of sign language motion using a dual-arm/hand humanoid robot,'' \emph{Intelligent Service Robotics}, vol.~9, no.~4, pp. 333--345, Oct. 2016.

\bibitem{hosseiniTeachingPersianSign2019c}
S.~R. Hosseini, A.~Taheri, A.~Meghdari, and M.~Alemi, ``Teaching {{Persian Sign Language}} to a {{Social Robot}} via the {{Learning}} from {{Demonstrations Approach}},'' in \emph{Social {{Robotics}}}, ser. Lecture {{Notes}} in {{Computer Science}}, M.~A. Salichs, S.~S. Ge, E.~I. Barakova, J.-J. Cabibihan, A.~R. Wagner, {\'A}.~{Castro-Gonz{\'a}lez}, and H.~He, Eds.\hskip 1em plus 0.5em minus 0.4em\relax {Cham}: {Springer International Publishing}, 2019, pp. 655--665.

\bibitem{gagoSequencetoSequenceNaturalLanguage2019}
J.~J. Gago, V.~Vasco, B.~{\L}ukawski, U.~Pattacini, V.~Tikhanoff, J.~G. Victores, and C.~Balaguer, ``Sequence-to-{{Sequence Natural Language}} to {{Humanoid Robot Sign Language}},'' in \emph{{{EUROSIM}} 2019 {{Abstract Volume}}}, 2019.

\bibitem{liangDynamicMovementPrimitive2021a}
Y.~Liang, W.~Li, Y.~Wang, R.~Xiong, Y.~Mao, and J.~Zhang, ``Dynamic {{Movement Primitive}} based {{Motion Retargeting}} for {{Dual-Arm Sign Language Motions}},'' in \emph{2021 {{IEEE International Conference}} on {{Robotics}} and {{Automation}} ({{ICRA}})}, May 2021, pp. 8195--8201.

\bibitem{un}
``International day of sign language - un,'' \url{https://www.un.org/en/observances/sign-languages-day}, 2022, accessed: 2022-08-23.

\bibitem{who}
``Deafness and hearing loss - who,'' \url{https://www.who.int/news-room/fact-sheets/detail/deafness-and-hearing-loss}, 2021, accessed: 2022-08-23.

\bibitem{zhangKinematicMotionRetargeting2022}
H.~Zhang, W.~Li, J.~Liu, Z.~Chen, Y.~Cui, Y.~Wang, and R.~Xiong, ``Kinematic {{Motion Retargeting}} via {{Neural Latent Optimization}} for {{Learning Sign Language}},'' \emph{IEEE Robotics and Automation Letters}, vol.~7, no.~2, pp. 4582--4589, Apr. 2022.

\bibitem{koseHumanoidRobotAssisted2011}
H.~Kose, R.~Yorganci, and I.~I. Itauma, ``Humanoid robot assisted interactive sign language tutoring game,'' in \emph{2011 {{IEEE International Conference}} on {{Robotics}} and {{Biomimetics}}}, Dec. 2011, pp. 2247--2248.

\bibitem{koseEvaluationRobotAssisted2012}
H.~Kose, R.~Yorganci, E.~H. Algan, and D.~S. Syrdal, ``Evaluation of the {{Robot Assisted Sign Language Tutoring Using Video-Based Studies}},'' \emph{International Journal of Social Robotics}, vol.~4, no.~3, pp. 273--283, Aug. 2012.

\bibitem{koseEffectEmbodimentSign2015}
H.~K{\"o}se, P.~Uluer, N.~Akal{\i}n, R.~Yorganc{\i}, A.~{\"O}zkul, and G.~Ince, ``The {{Effect}} of {{Embodiment}} in {{Sign Language Tutoring}} with {{Assistive Humanoid Robots}},'' \emph{International Journal of Social Robotics}, vol.~7, no.~4, pp. 537--548, Aug. 2015.

\bibitem{uluerNewRoboticPlatform2015}
P.~Uluer, N.~Akal{\i}n, and H.~K{\"o}se, ``A {{New Robotic Platform}} for {{Sign Language Tutoring}},'' \emph{International Journal of Social Robotics}, vol.~7, no.~5, pp. 571--585, Nov. 2015.

\bibitem{zakipourRASALowCostUpperTorso2016}
M.~Zakipour, A.~Meghdari, and M.~Alemi, ``{{RASA}}: {{A Low-Cost Upper-Torso Social Robot Acting}} as a {{Sign Language Teaching Assistant}},'' in \emph{Social {{Robotics}}}, ser. Lecture {{Notes}} in {{Computer Science}}, A.~Agah, J.-J. Cabibihan, A.~M. Howard, M.~A. Salichs, and H.~He, Eds.\hskip 1em plus 0.5em minus 0.4em\relax {Cham}: {Springer International Publishing}, 2016, pp. 630--639.

\bibitem{scassellatiTeachingLanguageDeaf2018}
B.~Scassellati, J.~Brawer, K.~Tsui, S.~Nasihati~Gilani, M.~Malzkuhn, B.~Manini, A.~Stone, G.~Kartheiser, A.~Merla, A.~Shapiro, D.~Traum, and L.-A. Petitto, ``Teaching {{Language}} to {{Deaf Infants}} with a {{Robot}} and a {{Virtual Human}},'' in \emph{Proceedings of the 2018 {{CHI Conference}} on {{Human Factors}} in {{Computing Systems}}}, ser. {{CHI}} '18.\hskip 1em plus 0.5em minus 0.4em\relax {New York, NY, USA}: {Association for Computing Machinery}, Apr. 2018, pp. 1--13.

\bibitem{zhiTeachingRobotSign2018}
D.~Zhi, T.~E.~A. {de Oliveira}, V.~P. da~Fonseca, and E.~M. Petriu, ``Teaching a {{Robot Sign Language}} using {{Vision-Based Hand Gesture Recognition}},'' in \emph{2018 {{IEEE International Conference}} on {{Computational Intelligence}} and {{Virtual Environments}} for {{Measurement Systems}} and {{Applications}} ({{CIVEMSA}})}, June 2018, pp. 1--6.

\bibitem{gagoSignLanguageRepresentation2019}
J.~J. Gago, J.~G. Victores, and C.~Balaguer, ``Sign {{Language Representation}} by {{TEO Humanoid Robot}}: {{End-User Interest}}, {{Comprehension}} and {{Satisfaction}},'' \emph{Electronics}, vol.~8, no.~1, p.~57, Jan. 2019.

\bibitem{meghdariDesignRealizationSign2019}
A.~Meghdari, M.~Alemi, M.~Zakipour, and S.~A. Kashanian, ``Design and {{Realization}} of a {{Sign Language Educational Humanoid Robot}},'' \emph{Journal of Intelligent \& Robotic Systems}, vol.~95, no.~1, pp. 3--17, July 2019.

\bibitem{luccioLearningSignLanguage2020}
F.~L. Luccio and D.~Gaspari, ``Learning {{Sign Language}} from a {{Sanbot Robot}},'' in \emph{Proceedings of the 6th {{EAI International Conference}} on {{Smart Objects}} and {{Technologies}} for {{Social Good}}}, ser. {{GoodTechs}} '20.\hskip 1em plus 0.5em minus 0.4em\relax {New York, NY, USA}: {Association for Computing Machinery}, Sept. 2020, pp. 138--143.

\bibitem{holdenPhasefunctionedNeuralNetworks2017}
D.~Holden, T.~Komura, and J.~Saito, ``Phase-functioned neural networks for character control,'' \emph{ACM Transactions on Graphics}, vol.~36, no.~4, pp. 42:1--42:13, July 2017.

\bibitem{pengDeepMimicExampleguidedDeep2018}
X.~B. Peng, P.~Abbeel, S.~Levine, and M.~{van de Panne}, ``{{DeepMimic}}: Example-guided deep reinforcement learning of physics-based character skills,'' \emph{ACM Transactions on Graphics}, vol.~37, no.~4, pp. 143:1--143:14, July 2018.

\bibitem{pengSFVReinforcementLearning2018}
X.~B. Peng, A.~Kanazawa, J.~Malik, P.~Abbeel, and S.~Levine, ``{{SFV}}: Reinforcement learning of physical skills from videos,'' \emph{ACM Transactions on Graphics}, vol.~37, no.~6, pp. 178:1--178:14, Dec. 2018.

\bibitem{wonScalableApproachControl2020}
J.~Won, D.~Gopinath, and J.~Hodgins, ``A scalable approach to control diverse behaviors for physically simulated characters,'' \emph{ACM Transactions on Graphics}, vol.~39, no.~4, pp. 33:33:1--33:33:12, Aug. 2020.

\bibitem{pengAMPAdversarialMotion2021}
X.~B. Peng, Z.~Ma, P.~Abbeel, S.~Levine, and A.~Kanazawa, ``{{AMP}}: Adversarial motion priors for stylized physics-based character control,'' \emph{ACM Transactions on Graphics}, vol.~40, no.~4, pp. 144:1--144:20, July 2021.

\bibitem{pengASELargeScaleReusable2022}
X.~B. Peng, Y.~Guo, L.~Halper, S.~Levine, and S.~Fidler, ``{{ASE}}: {{Large-Scale Reusable Adversarial Skill Embeddings}} for {{Physically Simulated Characters}},'' \emph{ACM Transactions on Graphics}, vol.~41, no.~4, pp. 1--17, July 2022.

\bibitem{schulmanProximalPolicyOptimization2017}
J.~Schulman, F.~Wolski, P.~Dhariwal, A.~Radford, and O.~Klimov, ``Proximal {{Policy Optimization Algorithms}},'' Aug. 2017.

\bibitem{haarnoja2018soft}
T.~Haarnoja, A.~Zhou, P.~Abbeel, and S.~Levine, ``Soft actor-critic: Off-policy maximum entropy deep reinforcement learning with a stochastic actor,'' \emph{International Conference on Machine Learning (ICML)}, 2018.

\bibitem{pybullet}
\BIBentryALTinterwordspacing
E.~Coumans and Y.~Bai, ``Pybullet, a python module for physics simulation for games, robotics and machine learning,'' 2016. [Online]. Available: \url{http://pybullet.org}
\BIBentrySTDinterwordspacing

\bibitem{kanazawaEndtoEndRecoveryHuman2018}
A.~Kanazawa, M.~J. Black, D.~W. Jacobs, and J.~Malik, ``End-to-{{End Recovery}} of {{Human Shape}} and {{Pose}},'' in \emph{2018 {{IEEE}}/{{CVF Conference}} on {{Computer Vision}} and {{Pattern Recognition}}}.\hskip 1em plus 0.5em minus 0.4em\relax {Salt Lake City, UT}: {IEEE}, June 2018, pp. 7122--7131.

\bibitem{rongFrankMocapMonocular3D2021}
Y.~Rong, T.~Shiratori, and H.~Joo, ``{{FrankMocap}}: {{A Monocular 3D Whole-Body Pose Estimation System}} via {{Regression}} and {{Integration}},'' Aug. 2021.

\bibitem{tavellaPhonologyRecognitionAmerican2022}
F.~Tavella, A.~Galata, and A.~Cangelosi, ``Phonology {{Recognition}} in {{American Sign Language}},'' in \emph{{{ICASSP}} 2022 - 2022 {{IEEE International Conference}} on {{Acoustics}}, {{Speech}} and {{Signal Processing}} ({{ICASSP}})}, May 2022, pp. 8452--8456.

\bibitem{tavellaWLASLLEXDatasetRecognising2022}
F.~Tavella, V.~Schlegel, M.~Romeo, A.~Galata, and A.~Cangelosi, ``{{WLASL-LEX}}: A {{Dataset}} for {{Recognising Phonological Properties}} in {{American Sign Language}},'' in \emph{Proceedings of the 60th {{Annual Meeting}} of the {{Association}} for {{Computational Linguistics}} ({{Volume}} 2: {{Short Papers}})}.\hskip 1em plus 0.5em minus 0.4em\relax {Dublin, Ireland}: {Association for Computational Linguistics}, May 2022, pp. 453--463.

\bibitem{frankimg}
\BIBentryALTinterwordspacing
Y.~Rong, T.~Shiratori, and H.~Joo, ``Frankmocap github repository,'' 2022. [Online]. Available: \url{https://github.com/facebookresearch/frankmocap/blob/main/docs/joint_order.md}
\BIBentrySTDinterwordspacing

\bibitem{suttonReinforcementLearningSecond2018}
R.~S. Sutton and A.~G. Barto, \emph{Reinforcement {{Learning}}, Second Edition: {{An Introduction}}}.\hskip 1em plus 0.5em minus 0.4em\relax {MIT Press}, Nov. 2018.

\bibitem{bellmanMarkovianDecisionProcess1957}
R.~Bellman, ``A {{Markovian Decision Process}},'' \emph{Journal of Mathematics and Mechanics}, vol.~6, no.~5, pp. 679--684, 1957.

\bibitem{caselliASLLEXLexicalDatabase2017}
N.~K. Caselli, Z.~S. Sehyr, A.~M. {Cohen-Goldberg}, and K.~Emmorey, ``{{ASL-LEX}}: {{A}} lexical database of {{American Sign Language}},'' \emph{Behavior Research Methods}, vol.~49, no.~2, pp. 784--801, Apr. 2017.

\end{thebibliography}

\end{document}